\documentclass{article}
\usepackage{tccml_iclr2025_conference}
\usepackage[utf8]{inputenc}
\usepackage{graphicx}
\usepackage{booktabs}
\usepackage{amsmath,amssymb,amsfonts}
\usepackage{algorithmic}
\usepackage{algorithm}
\usepackage{caption}
\usepackage{natbib}
\usepackage{url}
\usepackage{comment}
\title{SuoiAI: Building a Dataset for Aquatic Invertebrates in Vietnam}

\author{Tue Vo \\
  Nuoc Solutions \\
  \texttt{tue@nuoc.solutions}
  \And
  Lakshay Sharma \\
  Microsoft \\
  \texttt{lakshay.sharma@cims.nyu.edu}
  \And
  Tuan Dinh \\
  Nuoc Solutions \\
  \texttt{tuan@nuoc.solutions}
  \And
  Khuong Dinh \\
  University of Oslo \\
  \texttt{van.k.dinh@ibv.uio.no}
  \And
  Trang Nguyen \\
  Bowdoin College \\
  \texttt{t.nguyen@bowdoin.edu}
  \And
  Trung Phan \\
  Fulbright University Vietnam \\
  \texttt{trung.phan@fulbright.edu.vn}
  \And
  Minh Do \\
  Nuoc Solutions \\
  \texttt{minh@nuoc.solutions}
  \And
  Duong Vu \\
  Westerdijk Fungal Biodiversity Institute \\
  \texttt{d.vu@wi.knaw.nl}
}
\iclrfinalcopy
\begin{document}
\maketitle

\begin{abstract}
Understanding and monitoring aquatic biodiversity is critical for ecological health and conservation efforts. This paper proposes SuoiAI, an end-to-end pipeline for building a dataset of aquatic invertebrates in Vietnam and employing machine learning (ML) techniques for species classification. We outline the methods for data collection, annotation, and model training, focusing on reducing annotation effort through semi-supervised learning and leveraging state-of-the-art object detection and classification models. Our approach aims to overcome challenges such as data scarcity, fine-grained classification, and deployment in diverse environmental conditions.
\end{abstract}

\section{Introduction}
Aquatic invertebrates play a vital role in aquatic ecosystems and serve as indicators of water quality. For instance, streams are relatively small running surface water systems, yet they support an exceptionally high invertebrate biodiversity relative to their size. Stream invertebrates form a highly diverse group, encompassing insects, crustaceans, annelids, mollusks, and bivalves. These organisms play crucial roles in aquatic biogeochemical cycling and facilitate interactions between aquatic and terrestrial ecosystems. The lack of comprehensive datasets for these species, especially in biodiverse regions like Vietnam, poses significant challenges for ecological research and conservation. This paper proposes SuoiAI, a pipeline to build a robust dataset for aquatic invertebrates and leverage ML techniques for efficient species classification (Suoi means streams in Vietnamese). By addressing the data collection and annotation challenges, our work aims to facilitate scalable and accurate aquatic biodiversity monitoring in global hotspots like Vietnam.

\subsection{Problem Description}
Vietnam's aquatic ecosystems remain underrepresented in global biodiversity databases. Of over 100,000 freshwater invertebrate species described globally, only 2,000 have been identified in Vietnam, with 800 aquatic species lacking systematic documentation (\cite{thuaire2021assessing}). This significant gap limits our ability to assess aquatic health and understand climate change's impact on biodiversity. Without a comprehensive database, conservation efforts and policy decisions remain uninformed. Building Vietnam's first aquatic/ invertebrate database would address this foundational gap, enabling scalable and accurate biodiversity assessments. SuoiAI aims to leverage modern AI/ML techniques to overcome traditional field collection's limitations, delivering a transformative tool for biodiversity conservation and supporting global sustainability goals.

\subsection{Related Works}
There are a few datasets for aquatic invertebrates, most notably the iNaturalist Species  Classification and Detection Dataset (\cite{vanhorn2018inaturalistspeciesclassificationdetection}), the Benthic Macroinvertebrate Database from the Stroud Water Research Center\footnote{https://stroudcenter.org/macros/}, and the Freshwater Biological Traits Database from the Environmental Protection Agency (EPA) \footnote{https://www.epa.gov/risk/freshwater-biological-traits-database-traits}. However, many of these projects focus on American or European species and contain very little information for species from the tropical aquatic environment in Southeast Asia.

There have been a number of projects that aim to use automatic camera capture to study aquatic and marine invertebrates (\cite{heuschele2019imaging}, \cite{albini2024temporal}). In late 2024, Microsoft's AI for Good Lab also announced Project SPARROW (Solar-Powered Acoustic and Remote Recording Observation Watch)\footnote{https://blogs.microsoft.com/on-the-issues/2024/12/18/announcing-sparrow-a-breakthrough-ai-tool-to-measure-and-protect-earths-biodiversity-in-the-most-remote-places/}, which is an AI-powered edge computing solution designed to operate autonomously in remote corners of the planet to collect and process biodiversity data from sensors with edge GPUs and transmit via low-Earth orbit satellites directly to the cloud. SuoiAI can collaborate with Project SPARROW and other related projects to develop solar-powered aquatic cameras that can operate autonomously in the rainforests of Vietnam and Southeast Asia.

\section{Dataset Collection Setup}
\subsection{Data Acquisition}
Our solution proposes deploying underwater cameras in key aquatic ecosystems across Vietnam, including rivers, lakes, and coastal areas. The resolution of the cameras will be set to 1080p to balance image quality and storage requirements. Specimen density in the images will range between 1 to 5 specimens per image, minimizing occlusion and ensuring clear captures. Size of each invertebrate is about 10-50mm. Sampling will be conducted across various ecosystems to ensure diversity and generalizability of the dataset. Our initial field test sites include two important biological reserves of Vietnam: Cat Tiên national park and Cuc Phuong national park. The system's geographic coverage can expand to 135 sites nationwide, across latitudinal, altitudinal, and anthropogenic intensity gradients. Our automated systems are designed to generate around three million data points per site annually.

\subsection{Annotation Strategy}
To create a labeled dataset, we will begin with the manual labeling of a few hundred high-quality images for genus and species identification. Annotation tools like LabelImg and VIA will be utilized for bounding box and segmentation mask creation. Our annotators include biologists and scientists who are familiar with Vietnamese species. We will also look to include existing datasets of regional invertebrates to create the initial dataset.

Given the high cost of manual annotation, we propose several strategies to minimize effort. Initially, a teacher model, similar to \cite{chen2023mixedpseudolabelssemisupervised}, will be trained on manually labeled data to label additional images, which will then be used to iteratively train student models. This bootstrap labeling process reduces reliance on manual annotations. Few-shot and zero-shot learning have been explored to utilize pre-trained models like CLIP and Segment Anything Model (\cite{ravi2024sam2segmentimages}) for classification and segmentation of unseen insect and vertebrate species (\cite{feuer2024zeroshot}, \cite{Blair2021}, \cite{lu2024metric}, \cite{VILLON2021101320}). Alternative approaches such as computing clustering with vision embeddings will also help to group similar images to streamline targeted annotation (such as MorphoCluster by \cite{s20113060}). 

\begin{figure}[h!]
    \centering
    \includegraphics[width=0.8\textwidth]{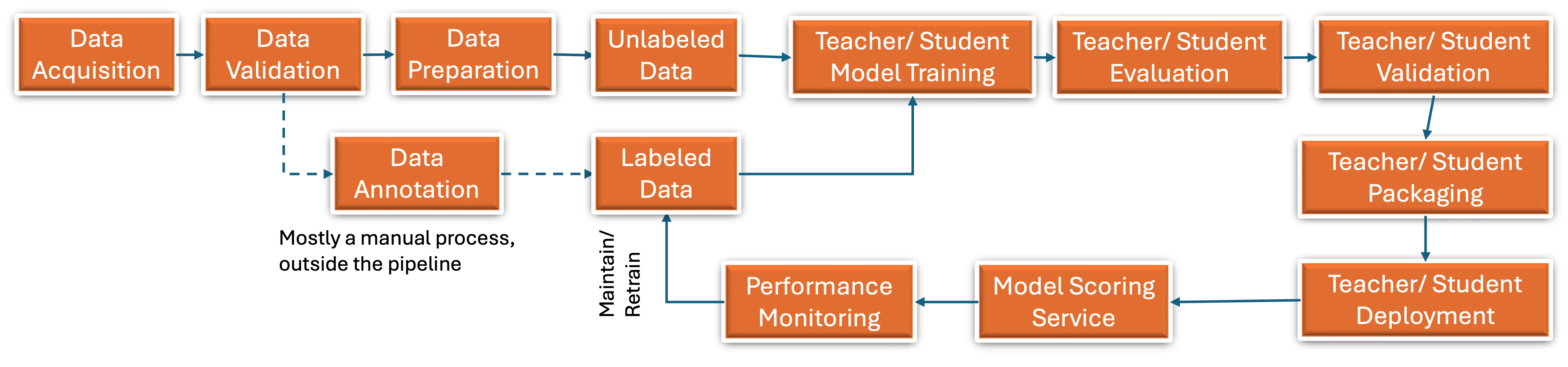}
    \caption{An end-to-end diagram of our pipeline}
    \label{fig:example-image}
\end{figure}

\section{Modeling Techniques}
\subsection{Baseline Object Detection}
We propose using conventional neural networks for object detection. Models like YOLO (\cite{redmon2016lookonceunifiedrealtime}) or Faster R-CNN (\cite{ren2016fasterrcnnrealtimeobject}) variants are older but small enough to be deployable on-device and therefore can serve as a reliable baseline. Models like SwinTransformer (\cite{liu2021swintransformer}) and DETR (\cite{carion2020endtoendobjectdetectiontransformers}) are newer and more accurate, but have more parameters and may be operable only in the cloud. However, these methods depend on large labeled datasets and require retraining for new species, which can be time-intensive.

\subsection{Training-Free and Semi-supervised Methods}
To complement supervised methods, training-free and semi-supervised approaches will be employed. Segment Anything Model (SAM) will be used for prompt-based segmentation with bounding boxes or points. While SAM is scalable and training-free, it may have limitations in accuracy for novel species. Semi-supervised learning will employ teacher-student paradigms to refine model predictions iteratively on unlabeled data, in a setup similar to \cite{chen2023mixedpseudolabelssemisupervised}. 

\subsection{Fine-grained Image Classification}
To address high intra-class similarity, fine-grained image classification techniques, similar to \cite{pham2024peeb}, will be utilized. Classification will be conducted at the genus and species levels (the later being more challenging), with approximately 20-50 genus-level classes and 100-200 species-level classes, with about 1000 labeled samples per class. These techniques will handle long-tail distributions and improve performance on rare classes (such as unknown species or genus). Other techniques to classify taxonomy classes with unknown species and genus include deep hierarchical Bayesian learning (such as \cite{bayesianlearning}).

\section{Practical Considerations}
Deployment challenges will be addressed through a dual-pronged approach: optimizing lightweight models (through quantization, model distillation, etc) for on-device processing in field conditions while leveraging cloud-based analysis for large-scale data processing. We will employ image super-resolution techniques to enhance low-quality images. The system must be robust in diverse aquatic conditions, addressing variables such as lighting, turbidity, and background noise. 

\section{Applications}
The proposed pipeline will serve as a powerful tool for in-situ biodiversity monitoring, identifying and classifying species while enabling the tracking of population dynamics and assessment of water quality and ecosystem health. Through its robust capabilities, this work will strengthen conservation efforts by discovering and documenting new species, and quantifying aquatic invertebrate biomass with precision. Furthermore, leveraging a curated label dataset, we aim to develop a Foundation Model for aquatic invertebrates, inspired by the Insect Foundation Model (\cite{nguyen2024insectfoundation}).

\section{Conclusion}
This paper outlines an integrated approach to building SuoiAI, a data pipeline for aquatic invertebrates in Vietnam, which aims to enhance biodiversity monitoring and conservation efforts in the region. Beyond Vietnam, SuoiAI holds promise for deployment in other tropical and equatorial regions, offering opportunities to uncover hidden biodiversity and ecological functions on a global scale.

\bibliography{Reference}

\end{document}